\documentclass[letterpaper]{article} 
\usepackage{aaai2026} 
\usepackage{times}  % DO NOT CHANGE THIS
\usepackage{helvet}  % DO NOT CHANGE THIS
\usepackage{courier}  % DO NOT CHANGE THIS
\usepackage[hyphens]{url}  % DO NOT CHANGE THIS
\usepackage{graphicx} % DO NOT CHANGE THIS
\usepackage{natbib} 
\usepackage{caption} 
\usepackage{algorithm}
\usepackage{algorithmic}
\usepackage{amssymb}
\usepackage{multirow}
\usepackage{siunitx}
\usepackage{graphicx}
\usepackage{amsmath}
\usepackage{makecell}
\usepackage[table,dvipsnames]{xcolor}
\usepackage{booktabs}
\usepackage{newfloat}
\usepackage{listings}
\usepackage{enumitem}
\usepackage{subcaption}
\DeclareCaptionStyle{ruled}{labelfont=normalfont,labelsep=colon,strut=off} % DO NOT CHANGE THIS
\floatstyle{ruled}
\newfloat{listing}{tb}{lst}{}
\floatname{listing}{Listing}
\title{VPN: Visual Prompt Navigation}
\author{
    Shuo Feng\textsuperscript{\rm 1,2}\equalcontrib\thanks{Work done as an Intern at Baidu Inc.},
    Zihan Wang\textsuperscript{\rm 3}\equalcontrib,
    Yuchen Li\textsuperscript{\rm 5},
    Rui Kong\textsuperscript{\rm 5}, Hengyi Cai\textsuperscript{\rm 5}, Shuaiqiang Wang\textsuperscript{\rm 5}, \\
    Gim Hee Lee\textsuperscript{\rm 3}, Piji Li\textsuperscript{\rm 1,2}\thanks{Corresponding author.}, Shuqiang Jiang\textsuperscript{\rm 4}
}
\affiliations{
\textsuperscript{\rm 1}College of Artificial Intelligence, Nanjing University of Aeronautics and Astronautics\\
\textsuperscript{\rm 2}The Key Laboratory of Brain-Machine Intelligence Technology, Ministry of Education\\
\textsuperscript{\rm 3}National University of Singapore
\textsuperscript{\rm 4}University of Chinese Academy of Sciences
\textsuperscript{\rm 5}Baidu Inc.\\
\{fengshuo,~pjli\}@nuaa.edu.cn, zihan.wang@u.nus.edu
}
\usepackage{bibentry}
\begin{document}

\maketitle

\begin{abstract}
While natural language is commonly used to guide embodied agents, the inherent ambiguity and verbosity of language often hinder the effectiveness of language-guided navigation in complex environments. To this end, we propose Visual Prompt Navigation (VPN), a novel paradigm that guides agents to navigate using only user-provided visual prompts within 2D top-view maps. This visual prompt primarily focuses on marking the visual navigation trajectory on a top-down view of a scene, offering intuitive and spatially grounded guidance without relying on language instructions. It is more friendly for non-expert users and reduces interpretive ambiguity. We build VPN tasks in both discrete and continuous navigation settings, constructing two new datasets, R2R-VP and R2R-CE-VP, by extending existing R2R and R2R-CE episodes with corresponding visual prompts. Furthermore, we introduce VPNet, a dedicated baseline network to handle the VPN tasks, with two data augmentation strategies: view-level augmentation (altering initial headings and prompt orientations) and trajectory-level augmentation (incorporating diverse trajectories from large-scale 3D scenes), to enhance navigation performance. Extensive experiments evaluate how visual prompt forms, top-view map formats, and data augmentation strategies affect the performance of visual prompt navigation. The code is available at https://github.com/farlit/VPN.
\end{abstract}

% Uncomment the following to link to your code, datasets, an extended version or similar.
% You must keep this block between (not within) the abstract and the main body of the paper.
% \begin{links}
%     \link{Code}{https://aaai.org/example/code}
%     \link{Datasets}{https://aaai.org/example/datasets}
%     \link{Extended version}{https://aaai.org/example/extended-version}
% \end{links}

\section{Introduction}

Developing robots that can navigate based on human instructions
has been a continuous research focus in artificial intelligence (AI) and robotics. Towards this goal, the research community has developed successive navigation paradigms spanning from PointGoal navigation (PointNav)~\cite{gupta2017cognitive} and ImageGoal navigation (ImageNav)~\cite{zhu2017target} to more complex formulations including object navigation (ObjectNav)~\cite{savva2017minos}, vision-and-language navigation (VLN)~\cite{anderson2018vision}. Despite significant progress in these visual navigation paradigms, they still have some limitations.
%Despite significant progress in these forms of visual navigation, they still have some limitations: 1) The inherent ambiguity of language instructions in ObjectNav and VLN makes perfectly precise navigation difficult to achieve. 2) In ImageNav, the absence of intermediate navigation cues poses challenges for the agent’s navigation. Moreover, requiring accurate images of the destination beforehand is unnatural and user-unfriendly.

\begin{figure}[!t]
    \hspace{-0.01\textwidth}
    \includegraphics[width=0.48\textwidth]{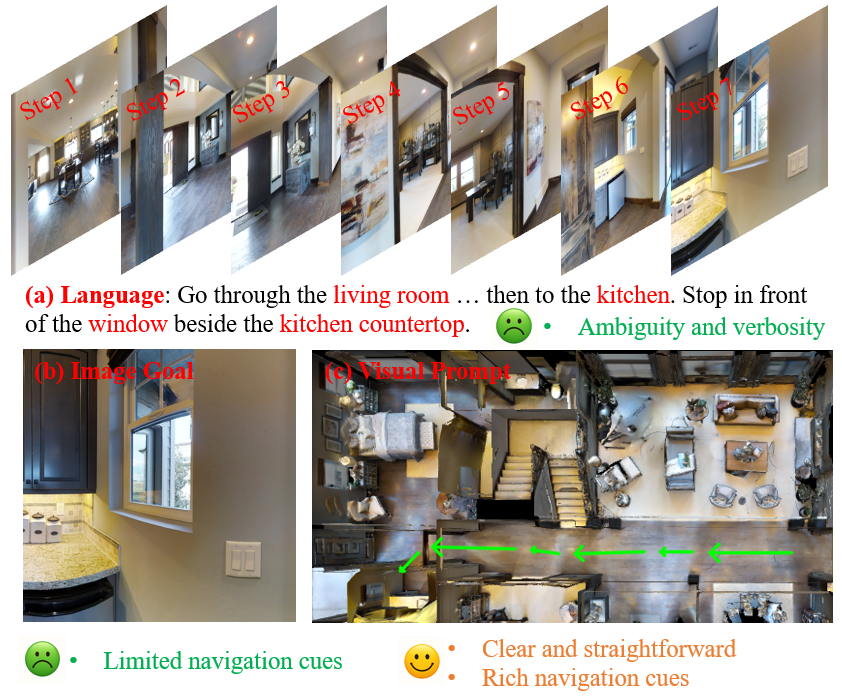} 
    \caption{Illustration of three types of visual navigation tasks. Compared to both language-based and image-goal instructions, visual prompt instructions provides clearer and more interpretable guidance.}
    \label{intro_small} 
\end{figure}

As Figure \ref{intro_small} illustrates, natural language often falls short in precisely specifying navigation trajectories. This is particularly evident when describing object positions, directional transitions, or distance relationships~\cite{chang2024survey}. Furthermore, the pursuit of linguistic precision through exhaustive description inevitably leads to prohibitive verbosity. This exposes an inherent dilemma in human-robot interaction: linguistic precision often comes at the cost of excessive verbosity~\cite{bonarini2020communication,wang2024large}. However, overly verbose language instructions often impose high demands on users and result in poor user experiences. In contrast to language instructions, image goals offer more direct and precise navigation targets. However, they suffer from two key limitations: 1) Lack of intermediate cues. The absence of intermediate navigation cues makes it difficult for agents to plan the correct route. 2) Impractical user demands. The requirement to pre-capture a large number of goal images as various destination candidates imposes impractical demands on usability.

Motivated by these findings, we propose a new navigation paradigm, called \textbf{Visual Prompt Navigation (VPN)}, which guides robots in 3D environments using user-provided visual prompts within 2D top-view maps for navigation. Visual prompt navigation aims to explore a more user-friendly interaction paradigm between users and navigation agents, providing navigation commands by simple and intuitive visual prompts. It offers several key benefits: 1) High user accessibility, allowing non-expert users to naturally specify navigation targets through clicking or drawing navigation trajectories on a visualized interface (e.g., top-view image). 2) Rich spatial information, as the top-view map with visual prompts inherently preserves complete spatial layouts and contains rich spatiotemporal trajectory descriptions, providing robots geometry-aware guidance, as illustrated in Figure~\ref{intro_small} (c). 3) High reusability, since such top-view maps can be acquired through aerial photography by drone or multi-view 3D scene reconstruction, and only need to be constructed once for repeated use across different navigation episodes. We believe VPN offers an efficient and user-friendly solution for robot navigation.

To verify the feasibility of using visual prompts as agent navigation guidance, we explore two navigation settings: the discrete VPN setting and the continuous VPN setting (VPN-CE), transferred from the VLN benchmarks: Room-to-Room (R2R)~\cite{anderson2018vision} and R2R-CE~\cite{krantz2020beyond} datasets, respectively. In both settings, we design visual prompts within 2D top-view maps, where users can simply mark a sequence of key navigation waypoints, which are then connected by arrows to indicate the expected navigation trajectory. Based on such visual prompts, we construct two new datasets: R2R with visual prompts (R2R-VP) and R2R-CE with visual prompts (R2R-CE-VP), which serve as standardized benchmarks for VPN tasks in discrete and continuous environments.

To tackle the VPN task, we propose a new method called \textbf{Visual Prompt Navigation Network (VPNet)}. In particular, we implement a discrete navigation setting of VPNet based on DUET~\cite{chen2022think}, and a continuous navigation setting based on ETPNav~\cite{an2024etpnav}, where the original instruction encoders in both models are replaced with a ViT-based~\cite{dosovitskiy2020image} encoder to encode top-view maps with visual prompts. Furthermore, we introduce two data augmentation strategies to enhance navigation performance: 1) View-level augmentation, which increases view diversity by randomly altering the agent's initial heading and rotating the visual prompt maps. 2) Trajectory-level augmentation, which increases trajectory diversity by incorporating additional trajectories from both MP3D~\cite{chang2017matterport3d} scenes and HM3D~\cite{ramakrishnan2021habitat} scenes. Besides, we validate how visual prompt forms, top-view map formats, and data augmentation strategies affect VPN performance.

In summary, we make the following contributions:
\begin{itemize}[leftmargin=1em]
\item We propose a new navigation task, Visual Prompt Navigation (VPN), which explores guiding the navigation agent based on visual prompts, and construct two corresponding datasets, R2R-VP and R2R-CE-VP, as standardized benchmarks in discrete and continuous settings.
\item We introduce VPNet, a baseline model designed for the VPN task, and incorporate two data augmentation strategies, view-level augmentation and trajectory-level augmentation, to enhance navigation performance.
\item Extensive experiments evaluate how visual prompt forms, top-view map formats, and data augmentation strategies affect the performance of visual prompt navigation.
\end{itemize}

\section{Related Work}

As VPN is a newly introduced task, we provide a brief overview of several closely related studies in the areas of visual navigation and visual prompt.

\noindent\textbf{Visual Navigation} requires the agent to utilize visual information to reach a specified destination. Based on the form of input instructions, it can be broadly categorized into five types: PointGoal navigation (PointNav)~\cite{anderson2018evaluation}, object navigation (ObjNav)~\cite{savva2017minos}, vision-and-language navigation (VLN)~\cite{anderson2018vision}, visual audio navigation (VAN)~\cite{chen2020soundspaces} and ImageGoal navigation (ImageNav)~\cite{zhu2017target}. Besides, FloNa~\cite{li2025flona} leverages floor plan outlines to guide agent navigation, but it heavily relies on accurate location information of both the agent and the destination, similar to PointNav. In PointNav, the agent is provided with the relative direction and distance to the goal. In contrast, our proposed VPN guides navigation solely via visual prompts within 2D top-view maps. This design can be seen as a hybrid of VLN and ImageNav: it combines the spatiotemporal trajectory descriptions characteristic of VLN with the intuitive visual presentation style of ImageNav.

\noindent\textbf{Visual Prompt}~\cite{wu2024visual} has recently emerged as a new paradigm that complements textual prompts by enabling more fine-grained, pixel-level instruction. Building on this idea, several studies have introduced symbolic elements, such as numbers, sketches, and arrows, to enhance natural language and improve the accuracy of Visual Question Answering (VQA)~\cite{yang2023dawn,cai2024vip}. Furthermore, recent works have explored the role of visual prompts in embodied AI (EAI). One line of work leverages generated visual prompts to improve agents' planning capabilities, including robotic manipulation~\cite{lee2024affordance,liu2024moka} and continuous VLN~\cite{chen2025affordances}. However, they still rely primarily on natural language input, with visual prompts serving only a supplementary role. Another line of work, represented solely by RoVI~\cite{li2025robotic}, uses hand-drawn symbolic representations as the sole form of instruction for robotic manipulation. While the above work has demonstrated the role of visual prompts in EAI, using visual prompts as the sole instruction input in visual navigation remains unexplored. Unlike the relatively static scenes in robotic manipulation, the scenes in visual navigation change continuously as the agent moves. Therefore, in this work, we are the first to adopt a compact scene representation using 2D top-view maps, where visual prompts are overlaid to serve as the sole guidance for agent navigation, without relying on any textual information. 

\section{VPN Dataset}
We transfer the VLN datasets Room-to-Room (R2R) and R2R-CE to VPN tasks by replacing language instructions with visual prompts in their navigation episodes, building two new datasets for VPN: R2R-VP and R2R-CE-VP, corresponding to the discrete and continuous environments, respectively. In this section, we detail the dataset construction process and provide an analysis of these datasets.

\subsection{Dataset Construction}

\begin{figure}[!t]
    \hspace{-0.01\textwidth}
    \includegraphics[width=0.48\textwidth]{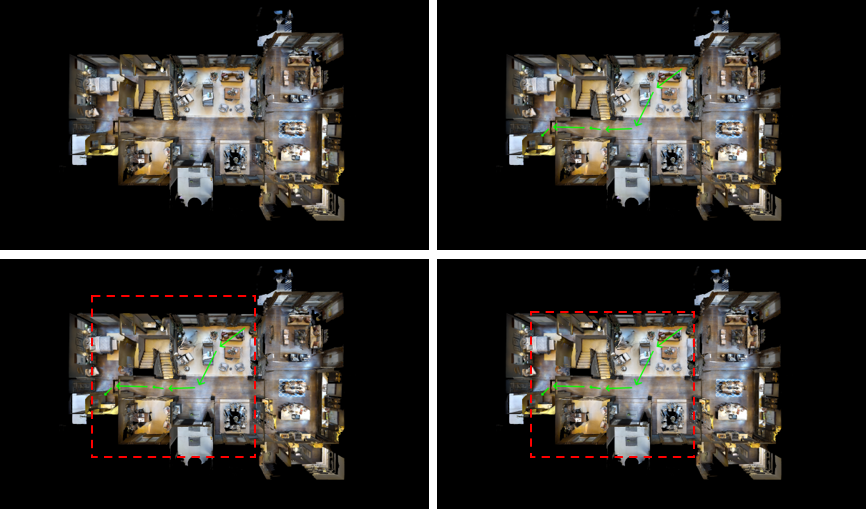} 
    \caption{Illustration of the process of constructing visual prompts on 2D top-view map. The four subfigures are labeled (a)–(d) from left to right, top to bottom.}
    \label{data_construct} 
\end{figure}

As Figure~\ref{data_construct} illustrates, we construct the R2R-VP and R2R-CE-VP datasets through a four-step process: a) For each episode in R2R and R2R-CE, we generate a 2D top-view map of the corresponding scene from 90 MP3D~\cite{chang2017matterport3d} scenes, following a procedure similar to LED~\cite{hahn2020you}. b) We then compute the pixel coordinates of all viewpoints on the top-view map and connect them sequentially with arrows to represent the agent’s navigation trajectory. c) Next, we apply a center-cropping operation to extract a square region centered on the trajectory. The side length is determined by the greater of the trajectory’s width or height, plus an additional 60-pixel margin, ensuring the full trajectory is included while maintaining spatial focus. d) Finally, to remove unnecessary black borders and tightly fit the visual prompts, we detect the bounding box of all non-zero pixels and crop accordingly. This process yields our two new datasets: R2R-VP and R2R-CE-VP. To further improve generalization and coverage, we enrich the training datasets by incorporating additional in-domain episodes from PREVALENT~\cite{hao2020towards} (based on MP3D scenes) and out-of-domain episodes from ScaleVLN~\cite{wang2023scaling} (based on HM3D~\cite{ramakrishnan2021habitat} scenes).

\subsection{Dataset Analysis}

\begin{table}
\small
\noindent\begin{minipage}[t]{1\columnwidth}%
\setlength{\tabcolsep}{1.15mm}
\centering
\begin{tabular}{c|c|cc|cc}
\toprule
\multirow{2}{*}{Dataset} & \multirow{2}{*}{Setting} & \multicolumn{2}{c|}{VLN}  & \multicolumn{2}{c}{VPN} \tabularnewline 

 & & Scenes & Episodes & Scenes & Episodes \tabularnewline \hline 
 
 \multirow{2}{*}{R2R} & Discrete & 61 & \num{4675} & 61 & \num{4638} \tabularnewline  
 & Continuous & 61 & \num{3603} & 60 & \num{3597} \tabularnewline \hline 
 
 \multirow{2}{*}{PRE} & Discrete & 60 & \num{178270} & 60 & \num{177134} \tabularnewline 
 & Continuous & - & - & 59 & \num{88694} \tabularnewline \hline 
 
 SCA & Discrete & 1289 & \num{4941710} & 523 & \num{1600945} 
 \tabularnewline 
\bottomrule
\end{tabular}
\end{minipage}
\caption{Statistics of training datasets for VLN and VPN tasks under discrete and continuous navigation settings. ``PRE'' denotes the PREVALENT dataset. ``SCA'' denotes the ScaleVLN dataset.}
\label{data_analysis1}
\end{table}

% \begin{table}
% \small
% \noindent\begin{minipage}[t]{1\columnwidth}%
% \tabcolsep=0.1cm
% \renewcommand{\arraystretch}{1.4}
% \centering
% \caption{Statistics of floor indexes in mp3d.}
% \label{data_analysis2}
% \begin{tabular}{c|c|c|c|c|c}\hline
%  Floor & 1 & 2 & 3 & 4 & 5 \tabularnewline \hline 
%  Number & 90 & 58 & 26 & 9 & 1 \tabularnewline \hline 
% \end{tabular}
% \end{minipage}
% \end{table}

Table~\ref{data_analysis1} summarizes the training dataset statistics for both VLN and VPN tasks under discrete and continuous navigation settings. It includes R2R, PREVALENT (constructed from MP3D scenes), and ScaleVLN (built from both HM3D and Gibson~\cite{xia2018gibson} scenes). In the ScaleVLN dataset used for VPN, we retain 523 scenes from HM3D for training. In general, certain entries under VPN contain fewer scenes or episodes compared to VLN, mainly due to two reasons: 1) Regarding scenes, some scenes are excluded due to poor reconstruction quality in their corresponding top-down maps, which makes them unsuitable for annotating reliable visual prompts. For MP3D, where each scene consists of multiple floors with separate top-down views, we manually exclude floors with poor reconstruction. For the HM3D scenes, we retain only the first floor, which generally has the best reconstruction quality. We apply the above strategies to transfer the VLN datasets to visual prompt navigation, including R2R for both continuous and discrete settings, as well as ScaleVLN and PREVALENT for the discrete setting. 2) Regarding episodes, some episodes are excluded because certain viewpoints cannot be mapped to MP3D’s 3D meshes~\cite{krantz2020beyond} under the continuous navigation setting. As a result, only valid viewpoints from R2R-CE are retained, and any episodes with invalid viewpoints in the PREVALENT dataset are discarded.

\section{VPNet}

\noindent\textbf{Problem Formulation.} In the discrete navigation setting of VPN, the environment is an undirected navigation connectivity graph $\mathcal{G} = (\mathcal{V}, \mathcal{E})$, where $\mathcal{V}$ denotes navigable nodes and $\mathcal{E}$ represents the edges indicating connectivity between nodes. Starting from an initial node and guided by visual prompts, the agent is required to explore the graph $\mathcal{G}$ and navigate to the target node. The visual prompt embeddings $\mathcal{P} = {\{vp_i\}}_{i=1}^m$ are encoded by a ViT-based~\cite{dosovitskiy2020image} encoder. At each time step $t$, the agent observes panoramic views $V_t=\{v_{t,i}\}_{i=1}^{n}$. The agent is aware of a few navigable views $\mathcal{N}(V_t)\subseteq V_t$, each corresponding to a neighboring node and its spatial coordinates.

VPN in continuous environments (VPN-CE) is established over Habitat~\cite{savva2019habitat} simulator. At each navigation step, we employ a pretrained waypoint predictor~\cite{hong2022bridging} to generate navigable waypoints in continuous environments, thereby aligning the task formulation with that of VPN in discrete environments.

\begin{figure}[ht]
  \centering
  \includegraphics[width=\linewidth]{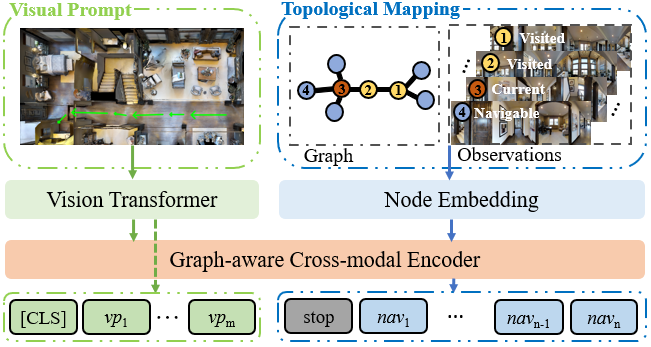}
  \caption{Illustration of VPNet. ``\textit{vp}\textsubscript{m}'' denotes token corresponding to visual prompts. ``stop'' indicates the token for the stop action, and ``\textit{nav}\textsubscript{n}'' represents token corresponding to navigable candidate.}
  \label{fig_method}
\end{figure}

\noindent\textbf{Overview.} We propose the Visual Prompt Navigation Network (VPNet) to address the newly introduced VPN task. Specifically, VPNet is based on DUET~\cite{chen2022think} in the discrete navigation setting and ETPNav~\cite{an2024etpnav} in the continuous navigation setting. The architecture of VPNet, as illustrated in Figure~\ref{fig_method}, comprises three main components: a ViT-based Visual Prompt Encoder, a Node Embedding module, and a Graph-aware Cross-modal Encoder. 

\subsection{Visual Prompt Processing}
\label{v2p}

To ensure consistent input size, all 2D top-view maps with visual prompts are resized to $224 \times 224$. To account for floor transitions caused by actions like ascending or descending stairs, we introduce an Order-Aware Floor Concatenation (OAFC), which concatenates visual prompt features while preserving their traversal order. Among 90 MP3D scenes, 32 have one floor, while the rest span multiple: 32 (2 floors), 17 (3), 8 (4), and 1 (5). In contrast, all 523 HM3D scenes used for training are single-floor. The final prompt feature extraction is defined as:
\begin{equation}
    \mathcal{P}_i^o =  \text{ViT}(\mathcal{P}_i)+b_i \tag{1} \label{prompt_order}
\end{equation}
\begin{equation}
    \mathcal{P} =  [\mathcal{P}_1^o,\mathcal{P}_2^o, \dots,\mathcal{P}_k^o] \tag{2} \label{agg_prompt}
\end{equation}
where $\mathcal{P}_i$ represents the $i$-th input visual prompt map, $b_i$ denotes the order embedding encoding the traversal sequence of floors, and $k$ denotes the total number of input visual prompt maps corresponding to the navigation trajectory. 

\subsection{Topological Mapping}
Since the agent has no prior knowledge of the environment graph $\mathcal{G}$, it incrementally constructs a topological graph based on observations gathered during navigation. At each time step $t$, the graph $\mathcal{G}_t = \{\mathcal{V}_t, \mathcal{E}_t \mid \mathcal{V}_t \subseteq \mathcal{V}, \mathcal{E}_t \subseteq \mathcal{E} \}$ comprises three types of nodes: the current node, visited nodes, and navigable nodes. The agent accesses panoramic views at current and visited nodes, and partial views of navigable nodes observed from current or previously visited locations. The current node $V_t$ and its neighboring unvisited nodes $\mathcal{N}(V_t)$ are added to $\mathcal{V}_{t-1}$ to update $\mathcal{V}_t$.

\noindent\textbf{Node Embedding.} Each node in the current and visited set is represented by aggregating the features of its panoramic views. Each view $W_{ij}'$ is computed as the sum of three components: the visual features $W_{ij}^v$, a direction embedding $W_{ij}^d$ that encodes the relative orientation between the view and the agent’s egocentric perspective, and a navigability type embedding $W_{ij}^n$ indicating whether the view leads to at least one adjacent, traversable node. The panoramic view encoder (a two-layer transformer) is then applied over all views of the node to capture inter-view contextual information, resulting in the contextual view representations $W_i$:
\begin{equation}
    W_{ij}' = \text{LN}(W_{ij}^v + W_{ij}^d + W_{ij}^n) \tag{3}
\end{equation}
\begin{equation}
    W_{i} = \text{SelfAttn}(W_{i}') \tag{4}
\end{equation}
where $i$ indexes the node in $\mathcal{V}_t$, $j$ indexes the view within that node, LN(·) and SelfAttn(·) denote layer normalization and the self-attention mechanism. For current and visited nodes, $W_i = (W{i1}, W_{i2}, \dots, W_{in})$ denotes the contextual representations of all $n$ panoramic views. For navigable nodes, which are only partially observed from visited or current locations, $W_i$ is constructed using the available subset of views with valid visual features $W_{ij}^v$. The current node is represented by averaging its local view features. Visited nodes retain their previously computed features. Navigable nodes are represented by averaging all partially observed views collected throughout the trajectory. In addition to view-based features, each node is further enriched with a step embedding $\mathcal{V}_{ti}^s$ encoding the agent’s visitation order, and a position embedding $\mathcal{V}_{ti}^p$ capturing spatial distance and orientation relative to other nodes. For unexplored nodes, the step embedding is set to 0. A special ``stop'' node is added to the graph to represent the stop action. The final node embedding is computed as:
\begin{equation}
    \mathcal{V}_{ti} = \mathcal{V}_{ti}^s + \mathcal{V}_{ti}^p + \frac{1}{n} \sum_{j=1}^{n} W_{ij} \tag{5}
\end{equation}
where $\mathcal{V}_{ti}$ is the $i$-th node in $\mathcal{V}_t = (\mathcal{V}_{t1}, \mathcal{V}_{t2}, \dots, \mathcal{V}_{tv})$, and $v$ is the total number of nodes at time $t$.

\subsection{Graph-Aware Navigation Policy}
\noindent\textbf{Graph-Aware Cross-Modal Encoder.} After obtaining the visual prompt embeddings $\mathcal{P}$ and node embeddings $\mathcal{V}_{t}$, we feed them into a multi-layer cross-modal graph transformer to model their interactions. Specifically, each transformer layer comprises a cross-attention layer to model the relations between visual prompts $\mathcal{P}$ and nodes $\mathcal{V}_{t}$, followed by a graph-aware self-attention (GASA) layer, which considers both visual similarity and spatial distances between nodes to obtain graph-aware node representations:
\begin{equation}
    X = \text{CrosAttn}(\mathcal{V}_{t}, \mathcal{P}) \tag{6}
\end{equation}
\begin{equation}
   \text{GASA}(X) = \text{Softmax}(\frac{XW_q(XW_k)^T}{\sqrt{d}}+EW_d)XW_v \tag{7}
\end{equation}
where CrosAttn(·) denotes the cross-attention mechanism, $X$ represents the stack of all node embeddings refined via cross-attention, $E$ is the pair-wise distance matrix obtained from $\mathcal{E}_t$, and $W_q$, $W_k$, $W_d$, $W_v$ are learnable matrices. The final visual-prompt aware representations of nodes are formulated as $\hat{v}=(\text{stop}, nav_1, nav_2, \dots, nav_{v})$, where $v$ indicates the total number of nodes in $\mathcal{V}_t$.

\begin{figure}[ht]
  \centering
  \includegraphics[width=\linewidth]{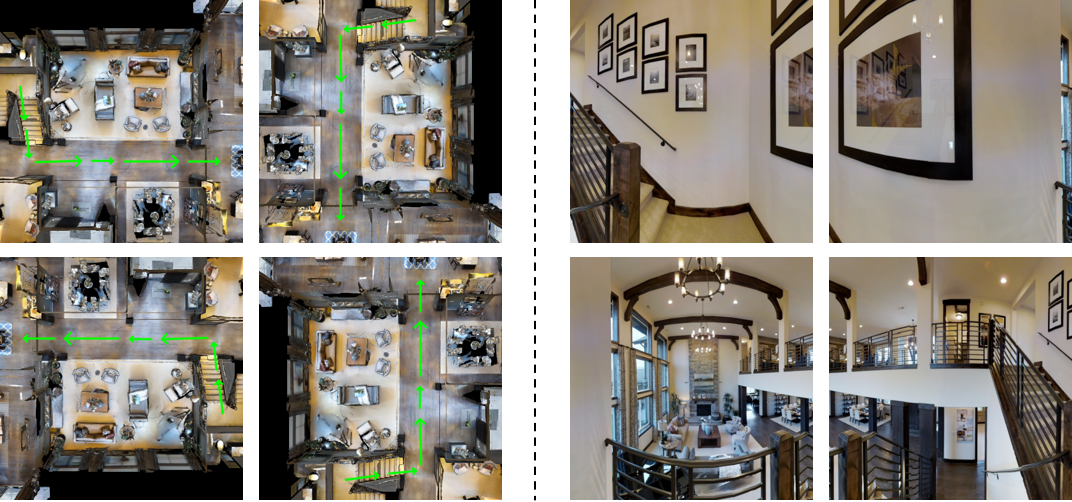}
  \caption{Illustration of prompt-view augmentation (left) and agent-view augmentation (right). The left side shows rotated 2D top-view maps with visual prompts at $0$, $\frac{\pi}{2}$, $\pi$, and $\frac{3\pi}{2}$. The right side shows the corresponding first-person observations when the agent starts with $0$, $\frac{\pi}{2}$, $\pi$, and $\frac{3\pi}{2}$.}
  \label{view_aug}
\end{figure}

\noindent\textbf{Global Action Prediction.} We apply a two-layer feed-forward network (FFN) to predict the navigation score $s_i$ for the node representation $\hat{v}_i$:
\begin{equation}
    s^g_i = \text{FFN}(\hat{v}_i) \tag{8}
\end{equation}
where it is worth noting that $s^g_0$ represents the stop score. To avoid redundant revisits and promote exploration, we mask the scores of previously visited nodes and the current node. The agent then chooses the highest-scoring candidate, either a navigable node or the ``stop'' node, and navigates to it via the shortest path in the topological graph $\mathcal{G}_t$.

\subsection{Data Augmentation}
\label{data_aug}
To further enhance navigation performance, we introduce two data augmentation strategies. 

\noindent\textbf{Trajectory-Level Augmentation.} To enhance trajectory diversity, we introduce trajectory-level augmentation by incorporating additional navigation episodes from in-domain and out-of-domain environments. Here, in-domain refers to environments included in the R2R training set, while out-of-domain refers to those that are not. Specifically, we include in-domain trajectories from the PREVALENT dataset~\cite{hao2020towards}, which are based on 60 MP3D~\cite{chang2017matterport3d} scenes (Chang et al., 2017), as well as out-of-domain trajectories from the ScaleVLN dataset~\cite{wang2023scaling}, which are constructed using 523 HM3D~\cite{ramakrishnan2021habitat} scenes. This augmentation strategy enriches the training data with diverse navigation paths, thereby expanding coverage across different scene types and improving the agent’s robustness to domain shifts.

\noindent\textbf{View-Level Augmentation.} As shown in Figure~\ref{view_aug}, we introduce two view-level augmentation strategies: 1) \textit{Prompt-view augmentation} diversifies visual prompts by randomly rotating the 2D top-view maps by $0$, $\frac{\pi}{2}$, $\pi$, or $\frac{3\pi}{2}$. 2) \textit{Agent-view augmentation} enhances the diversity of initial observations by randomly sampling the agent’s heading from $[0, 2\pi)$. Agent-view augmentation is particularly suitable for VPN tasks, where the initial heading is independent of the visual prompt. In contrast, for VLN tasks, instructions may implicitly specify the agent’s initial orientation, making such augmentation less applicable.

\subsection{Training}
Behavior cloning (BC) $\psi$ often suffers from distribution shifts between training and testing phases. To mitigate this issue, we fine-tune the VPNet using supervision from a pseudo interactive demonstrator $\psi^{*}$, following the DAgger algorithm~\cite{ross2011reduction}. The final loss combines Behavior cloning and DAgger supervision:
\begin{gather}
    \mathcal{L}_{\text{BC}} = \sum_{t=1}^{T} -\log  \, p(a^{\psi}_t|\mathcal{P}, \mathcal{G}_t) \tag{9}    \\
    \mathcal{L}_{\text{DAG}} = \sum_{t=1}^{T} -\log \, p(a^{\psi^{*}}_t|\mathcal{P}, \mathcal{G}_t^{*}) \tag{10}
\end{gather}
where $a^{\psi}_t$ denotes the ground truth action, $a^{\psi^{*}}_t$ represents the pseudo label, defined as the next node on the shortest path to the destination, computed over the partial graph $\mathcal{G}_t^{*}$ constructed by the agent via on-policy action sampling. The overall loss is defined as $\mathcal{L}=\lambda\mathcal{L}_{\text{BC}}+(1-\lambda)\mathcal{L}_{\text{DAG}}$, where $\lambda$ is a weighting factor that balances the two objectives.

\section{Experiments}
\subsection{Datasets and Evaluation Metrics}
We train our model using visual prompt maps corresponding to trajectories selected from the training split of R2R~\cite{anderson2018vision} and R2R-CE~\cite{krantz2020beyond}, PREVALENT~\cite{hao2020towards}, and ScaleVLN~\cite{wang2023scaling} datasets. For details on the validation and test splits, please refer to the appendix.

We adopt the same evaluation metrics in VLN to assess navigation performance in VPN, i.e., Trajectory Length (TL): the average path length in meters; Navigation Error (NE), the average distance in meters between the agent’s final location and the target location; Success Rate (SR), the percentage of paths with NE less than 3 meters; Oracle Success Rate (OSR), the success rate assuming an oracle stopping policy; and SR Penalized by Path Length (SPL).

\subsection{Implementation Details}
\noindent\textbf{Features.} We adopt the ViT-B/16~\cite{dosovitskiy2020image}, pretrained on ImageNet~\cite{deng2009imagenet}, to extract the panoramic view features $W_{ij}^v$ for each view in panoramas and to encode the visual prompt map $\mathcal{P}_i$ for each navigation episode across all datasets.

\noindent\textbf{Model Architecture.} We use 12, 2, 4, and 4 transformer layers for the visual prompt encoder (ViT-B/16), panoramic view encoder, graph-aware cross-modal encoder, and local cross-modal encoder (see the appendix for details), respectively, all with hidden size 768. VPNet includes both cross-modal encoders in discrete environments, but only the graph-aware encoder is used in continuous environments. We set $\lambda$ to 0.5 for both VPN and VPN-CE.

\noindent\textbf{Training Details.} For the VPN task, we train VPNet on the R2R-VP dataset along with additional episodes extracted from PREVALENT and ScaleVLN. VPNet is trained for 400k iterations on a single NVIDIA RTX A5000 GPU using a batch size of 10 and a learning rate of 1.5e-5. 
% The best checkpoint is selected based on the SPL on the unseen validation split of R2R-VP. 
For the VPN-CE task, VPNet is trained on R2R-CE-VP and episodes extracted from PREVALENT, using a total batch size of 16 across two NVIDIA RTX A5000 GPUs for 400k iterations, with a learning rate of 1e-5. 
% The best checkpoint is selected based on the SPL on the unseen validation split of R2R-CE-VP.

\begin{table*}[ht]
\centering
\setlength{\tabcolsep}{1.31mm}
\definecolor{Gray}{gray}{0.8}
\begin{tabular}{lcccc  cccc 
cccc }
\toprule
\multicolumn{1}{c}{\multirow{2}{*}{Methods}} & \multicolumn{4}{c}{R2R/R2R-VP Val Seen} & \multicolumn{4}{c}{R2R/R2R-VP Val Unseen} & \multicolumn{4}{c}{R2R/R2R-VP Test Unseen} \\
\cmidrule(lr){2-5} \cmidrule(lr){6-9} \cmidrule(lr){10-13} 
\multicolumn{1}{c}{} & 
\multicolumn{1}{c}{NE$\downarrow$} & \multicolumn{1}{c}{OSR$\uparrow$} & \multicolumn{1}{c}{SR$\uparrow$} & \multicolumn{1}{c}{SPL$\uparrow$} & 
\multicolumn{1}{c}{NE$\downarrow$} & \multicolumn{1}{c}{OSR$\uparrow$} & \multicolumn{1}{c}{SR$\uparrow$} & \multicolumn{1}{c}{SPL$\uparrow$} & 
\multicolumn{1}{c}{NE$\downarrow$} & \multicolumn{1}{c}{OSR$\uparrow$} & \multicolumn{1}{c}{SR$\uparrow$} & \multicolumn{1}{c}{SPL$\uparrow$} \\ 

\midrule
\rowcolor{Cerulean!20}
\multicolumn{13}{l}{\emph{Vision-and-Language Navigation Evaluated on R2R in Discrete Environments: }} \\

Seq2Seq~\cite{anderson2018vision}
& 6.01 & 53 & 39 & -- 
& 7.81 & 28 & 21 & -- 
& 7.85 & 27 & 20 & -- \\
RCM~\cite{wang2019reinforced}
& 3.53 & 75 & 67 & -- 
& 6.09 & 50 & 43 & -- 
& 6.12 & 50 & 43 & 38 \\
PREVALENT~\cite{hao2020towards}
& 3.67 & -- & 69 & 65
& 4.71 & -- & 58 & 53
& 5.30 & 61 & 54 & 51 \\
VLN$\circlearrowright$BERT~\cite{hong2020recurrent} 
& 2.90 & -- & 72 & 68
& 3.93 & -- & 63 & 57
& 4.09 & 70 & 63 & 57 \\
HAMT~\cite{chen2021history} 
& 2.51 & -- & 76 & 72
& 2.29 & -- & 66 & 61
& 3.93 & 72 & 65 & 60 \\
DUET~\cite{chen2022think} 
& 2.28 & 86 & 79 & 73 
& 3.31 & 81 & 72 & 60
& 3.65 & 76 & 69 & 59 \\
BEVBert~\cite{an2022bevbert} 
& 2.17 & 88 & 81 & 74
& 2.81 & 84 & 75 & 64
& 3.13 & 81 & 73 & 62 \\
GridMM~\cite{wang2023gridmm} 
& - & - & - & -
& 2.83 & - & 75 & 64
& 3.35 & - & 73 & 62 \\
Lily~\cite{lin2023learning} 
& - & - & - & -
& 2.90 & - & 74 & 62
& 3.44 & - & 72 & 60 \\
ScaleVLN~\cite{wang2023scaling} 
& 2.12 & 87 & 81 & 75
& 2.09 & 88 & 81 & 70
& 2.27 & 86 & 80 & 70 \\
NaviLLM~\cite{zheng2024towards} 
& - & - & - & -
& 3.51 & - & 67 & 59
& 3.71 & - & 68 & 60 \\
NavGPT-2~\cite{zhou2024navgpt} 
& 1.78 & 88 & 81 & 75
& 3.13 & 81 & 72 & 61
& 3.35 & 78 & 71 & 60 \\
VER~\cite{liu2024volumetric} 
& - & - & - & -
& 2.80 & - & 76 & 65
& 2.74 & - & 76 & 66 \\
MAGIC~\cite{wang2024magic} 
& 1.73 & 89 & 84 & 80
& 2.22 & 86 & 79 & 70
& 2.75 & 82 & 77 & 69 \\
SRDF~\cite{wang2024bootstrapping} 
& - & - & - & -
& 1.62 & 90 & 86 & 79
& 1.82 & 89 & 85 & 78 \\
DUET (R2R+PRE)
& 2.28 & 86 & 79 & 73
& 3.31 & 81 & 72 & 60
& 3.65 & 76 & 69 & 59 \\ 
DUET (R2R+PRE+SCA)
& 2.12 & 87 & 81 & 75
& 2.09 & 88 & 81 & 70
& 2.27 & 86 & 80 & 70 \\

\midrule
\rowcolor{Cerulean!20}
\multicolumn{13}{l}{\emph{Visual Prompt Navigation Evaluated on R2R-VP in Discrete Environments: }} \\

VPNet (R2R)
 & 3.37 & 72.41 & 66.69 & 62.13
 & 5.43 & 59.53 & 51.23 & 43.47
 & 5.11 & 62.33 & 52.40 & 42.87 \\

VPNet (R2R+PRE)
& \textbf{0.05} & \textbf{100} & \textbf{100} & \textbf{99.77}
& \underline{2.18} & \underline{76.76} & \underline{65.92} & \underline{56.17}
& \underline{1.94} & \underline{78.21} & \underline{66.38} & \underline{56.26} \\

VPNet (R2R+PRE+SCA)
& \underline{0.14} & \underline{99.41} & \underline{99.41} & \underline{99.08}
& \textbf{0.48} & \textbf{97.45} & \textbf{96.68} & \textbf{94.84}
& \textbf{0.31} & \textbf{98.56} & \textbf{97.56} & \textbf{94.60}\\
\bottomrule
\end{tabular}
\caption{Performance comparison between VPNet and existing VLN methods in discrete environments. ``PRE'' denotes the PREVALENT dataset. ``SCA'' denotes the ScaleVLN dataset. Best results are marked in bold and second best are underlined.}
\label{main_results1}
\end{table*}

\begin{table}[H]
\centering
\definecolor{Gray}{gray}{0.8}
\setlength{\tabcolsep}{1.3mm}
\scriptsize
\begin{tabular}{lcccccc}
\toprule
\multicolumn{1}{c}{\multirow{2}{*}{Methods}} & \multicolumn{3}{c}{Val Seen} & \multicolumn{3}{c}{Val Unseen}  \\
\cmidrule(lr){2-4} \cmidrule(lr){5-7} 
\multicolumn{1}{c}{} & \multicolumn{1}{c}{NE$\downarrow$} & \multicolumn{1}{c}{SR$\uparrow$} & \multicolumn{1}{c}{SPL$\uparrow$} & \multicolumn{1}{c}{NE$\downarrow$} & \multicolumn{1}{c}{SR$\uparrow$} & \multicolumn{1}{c}{SPL$\uparrow$} \\ 

\midrule
\rowcolor{Cerulean!20}
\multicolumn{7}{l}{\emph{VLN-CE Evaluated on R2R-CE: }} \\

Seq2Seq~\cite{krantz2020beyond} & 7.12 & 37 & 35 & 7.37 & 32 & 30 \\
CMTP~\cite{chen2021topological} & 7.10 & 36 & 31 & 7.90 & 26 & 23 \\
WPN~\cite{krantz2021waypoint} & 5.48 & 46 & 43 & 6.31 & 36 & 34 \\
LAW~\cite{raychaudhuri2021language} & 6.35 & 40 & 37 & 6.83 & 35 & 31 \\
CM2~\cite{georgakis2022cross} & 6.10 & 43 & 35 & 7.02 & 34 & 28 \\
WS-MGMap~\cite{chen2022weakly} & 5.65 & 47 & 43 & 6.28 & 39 & 34 \\
Sim2Sim~\cite{krantz2022sim} & 4.67 & 52 & 44 & 6.07 & 43 & 36 \\
CMA~\cite{hong2022bridging} & 5.20 & 51 & 45 & 6.20 & 41 & 36 \\
GridMM~\cite{wang2023gridmm} & 4.21 & 59 & 51 & 5.11 & 49 & 41 \\
ETPNav~\cite{an2024etpnav} & 3.95 & 66 & 59 & 4.71 & 57 & 49 \\
ERG~\cite{wang2024graph} & 5.04 & 46 & 42 & 6.20 & 39 & 35 \\
NaVILA~\cite{cheng2024navila} & - & - & - & 5.22 & 54 & 49 \\
ETPNav (R2R+PRE) & 3.95 & 66 & 59 & 4.71 & 57 & 49 \\

\midrule
\rowcolor{Cerulean!20}
\multicolumn{7}{l}{\emph{VPN-CE Evaluated on R2R-CE-VP: }} \\
VPNet (R2R) & \underline{4.75} & \underline{61.24} & \underline{55.88} & \underline{6.14} & \underline{42.09} & \underline{36.16} \\
VPNet (R2R+PRE) & \textbf{2.30} & \textbf{84.11} & \textbf{78.15} & \textbf{6.10} & \textbf{47.96} & \textbf{41.51} \\
\bottomrule
\end{tabular}
\caption{Performance comparison between VPNet and existing VLN methods in continuous environments.}
\label{main_results2}
\end{table}

\subsection{Main Results}
\label{main_result}
Tables~\ref{main_results1} and~\ref{main_results2} compare VPNet in VPN with prior VLN methods and highlight the benefits of trajectory-level data augmentation. Adding PREVALENT trajectories notably improves performance on both val seen and unseen splits, with VPNet achieving 100\% on SR in the discrete environment and 84.11\% in the continuous environment (val seen). Further incorporating trajectories from ScaleVLN leads to 96.68\% on SR in the discrete environment (val unseen). Notably, as shown in Table~\ref{main_results1}, VPNet in the VPN setting outperforms DUET in VLN while using only \num{1600945} trajectories from ScaleVLN, compared to DUET’s \num{4941710}, as reported in Table~\ref{data_analysis1}. These results highlight the data efficiency of scaling within the VPN framework, which refers to its ability to achieve strong performance with less training data, and further demonstrate the effectiveness of visual prompts in guiding robot navigation.

\subsection{Ablation Study}

\noindent\textbf{1) The Effect of Different Types of Visual Prompts.} As shown in row (e) of Table~\ref{abalation_vps}, the visual prompt consisting solely of lines outperforms that in row (d), which combines both lines and text. This is because the lines provide enough spatiotemporal guidance, while the text may obscure some details around the navigation waypoints. We attribute the poor performance of row (a) to the absence of trajectory-centered cropping, which causes episodes from the same scene to share identical top-view maps despite following different visual trajectories. This not only makes the VPNet prone to overfitting on the limited and unvarying top-view maps but also leads to underfitting of the crucial visual trajectory information. Interestingly, row (b), which only uses cropped top-view maps without including visual trajectories, 

\begin{figure}[ht]
  \centering
  \includegraphics[width=\linewidth]{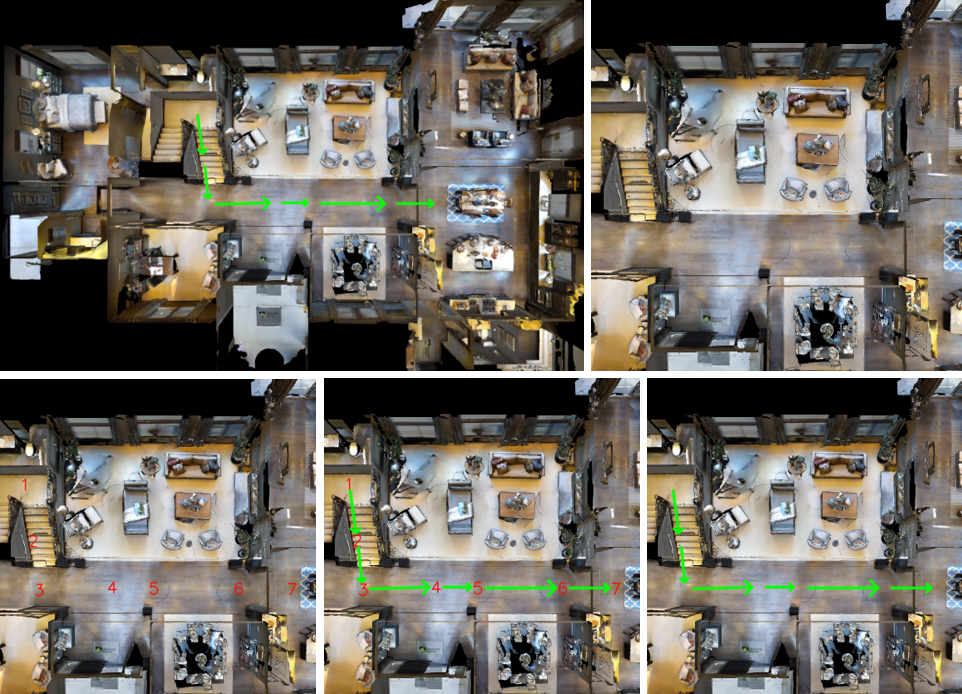}
  \caption{Illustration of different 2D top-view maps with visual prompts. The six subfigures are labeled (a)–(e) from left to right, top to bottom.}
  \label{vp_compare}
\end{figure}

\begin{table}[ht]
\centering
\scriptsize
\setlength{\tabcolsep}{3.23mm}
\begin{tabular}{c|ccc|ccc}
\toprule
\multicolumn{1}{c|}{\multirow{2}{*}{TVP}} & \multicolumn{3}{c|}{Val Seen} & \multicolumn{3}{c}{Val Unseen}  \\
\cline{2-7}
\multicolumn{1}{c|}{} & \multicolumn{1}{c}{NE$\downarrow$} & \multicolumn{1}{c}{SR$\uparrow$} & \multicolumn{1}{c|}{SPL$\uparrow$} & \multicolumn{1}{c}{NE$\downarrow$} & \multicolumn{1}{c}{SR$\uparrow$} & \multicolumn{1}{c}{SPL$\uparrow$}  \\ \hline
(a) & 6.86 & 31.68 & 28.63 & 7.07 & 33.94 & 29.36 \\
(b) & 1.47 & 83.56 & 78.64 & 5.51 & 45.83 & 36.78 \\
(c) & 1.04 & 88.29 & 83.37 & 5.13 & 50.18 & 40.54 \\
(d) & \underline{0.79} & \underline{95.74} & \underline{91.93} & \underline{3.74} & \underline{65.36} & \underline{54.81} \\
(e) & \textbf{0.05} & \textbf{100} & \textbf{99.77} & \textbf{2.18} & \textbf{65.92} & \textbf{56.17} \\
\bottomrule
\end{tabular}
\caption{Ablation study results on different types of visual prompts in discrete environments. ``TVP'' denotes the types of visual prompts, corresponding to subfigures in Figure~\ref{vp_compare}.}
\label{abalation_vps}
\end{table}

\begin{table}[H]
\centering
\scriptsize
\setlength{\tabcolsep}{3.5mm}
\begin{tabular}{c|ccccc}
\toprule
VLDA & 
\multicolumn{1}{c}{TL$\downarrow$} & \multicolumn{1}{c}{NE$\downarrow$} & \multicolumn{1}{c}{OSR$\uparrow$} & \multicolumn{1}{c}{SR$\uparrow$} & \multicolumn{1}{c}{SPL$\uparrow$}  \\ \hline
None & 11.47 & 1.69 & 90.75 & 86.33 & 82.92 \\
Agent view & 12.14 & 1.48 & 92.56 & 88.18 & 85.02 \\
Prompt view & 10.04 & \underline{0.62} & \underline{97.28} & \underline{96.41} & \underline{94.37} \\
Both & 9.92 & \textbf{0.48} & \textbf{97.45} & \textbf{96.68} & \textbf{94.84} \\
\bottomrule
\end{tabular}
\caption{Ablation study results on different combinations of view-level data augmentation (VLDA).}
\label{abalation_view_augs}
\end{table}
\noindent still achieves decent performance. We speculate that the navigation model may still be able to infer the approximate destination area from the cropped top-view map (similar to the image goal in the ImageNav task). However, due to the absence of explicit guidance from visual trajectories, its overall performance remains limited.

\noindent\textbf{2) The Effect of View-Level Data Augmentation.} As shown in Table~\ref{abalation_view_augs}, it is evident that both Prompt-view and Agent-view augmentations (Section~\ref{data_aug}) lead to improvements in performance. Notably, rotating the visual prompt yields significantly greater gains compared to randomly altering the agent's initial heading. This phenomenon remains consistent with our findings in Section~\ref{main_result}  that increasing the scale and diversity of top-view map data contributes significantly to improved navigation performance.

\begin{figure}[ht]
  \centering
  \includegraphics[width=\linewidth]{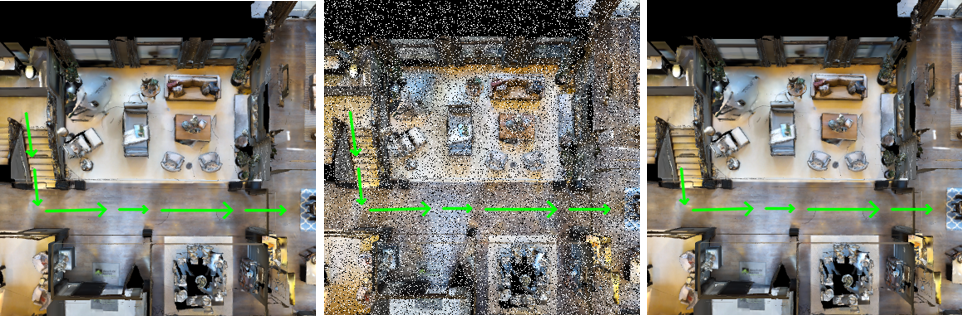}
  \caption{Illustration of visual prompt maps with different types of noise. From left to right, subfigures (a)–(c) represent: (a) the original visual prompt map, (b) the map with 20\% of its pixels replaced by salt-and-pepper noise, and (c) the map with the first-step visual cue removed.}
  \label{vp_noise}
\end{figure}

\begin{table}[ht]
\centering
\scriptsize
\setlength{\tabcolsep}{3.55mm}
\begin{tabular}{c|ccccc}
\toprule
Noise type & 
\multicolumn{1}{c}{TL$\downarrow$} & \multicolumn{1}{c}{NE$\downarrow$} & \multicolumn{1}{c}{OSR$\uparrow$} & \multicolumn{1}{c}{SR$\uparrow$} & \multicolumn{1}{c}{SPL$\uparrow$}  \\ \hline
(a) & 9.92 & \textbf{0.48} & \textbf{97.45} & \textbf{96.68} & \textbf{94.84} \\
(b) & 11.73 & \underline{1.24} & \underline{93.53} & \underline{90.34} & \underline{85.40} \\
(c) & 12.48 & 2.13 & 89.31 & 85.97 & 74.98 \\
\bottomrule
\end{tabular}
\caption{Evaluation of VPNet under different noise types corresponding to subfigures (a)–(c) of Figure~\ref{vp_noise}.}
\label{abalation_noise}
\end{table}

\begin{table}[H]
\centering
\scriptsize
\setlength{\tabcolsep}{3.2mm}
\begin{tabular}{c|ccccc}
\toprule
Encoder setting & 
\multicolumn{1}{c}{TL$\downarrow$} & \multicolumn{1}{c}{NE$\downarrow$} & \multicolumn{1}{c}{OSR$\uparrow$} & \multicolumn{1}{c}{SR$\uparrow$} & \multicolumn{1}{c}{SPL$\uparrow$}  \\ \hline
ViT-1k\&ViT-1k & 9.78 & 0.50 & 96.93 & 96.17 & \underline{94.60} \\
ViT-21k\&ViT-1k & 9.75 & \textbf{0.45} & \underline{97.13} & \underline{96.36} & 94.37 \\
ViT-21k\&ViT-21k & 9.92 & \underline{0.48} & \textbf{97.45} & \textbf{96.68} & \textbf{94.84} \\
\bottomrule
\end{tabular}
\caption{Ablation on encoder pairs: the left of ``\&'' encodes visual prompts, the right encodes scene observations. ``ViT-21k'' is pretrained on ImageNet-21k, ``ViT-1k'' is further fine-tuned from ``ViT-21k'' on its subset, ImageNet-1k.}
\label{abalation_vits}
\end{table}

\noindent\textbf{3) Noise Robustness Analysis of VPNet.} As shown in Table~\ref{vp_noise}, we evaluate VPNet under different types of noise corresponding to subfigures (a)–(c) in Figure~\ref{vp_noise}, simulating challenges such as low-quality top-view map reconstructions and inaccuracies in user-provided visual prompts. Although performance drops under these perturbations, VPNet, trained on clean data, still maintains strong results. This demonstrates its robustness to reconstruction-induced noise and user-induced inaccuracies.

\noindent\textbf{4) The Effect of Different Encoder Settings.} As Table~\ref{abalation_vits} illustrates, using ``ViT-21k'' for both visual prompts and scene observations yields the best performance. Despite being fine-tuned from ``ViT-21k'', ``ViT-1k'' is trained on a smaller set with fewer categories, which narrows the semantic space and limits generalization. This suggests that broader pretraining benefits visual understanding.

\section{Conclusion and Acknowledgements}
We present Visual Prompt Navigation using 2D top-view prompts to guide agents. We build benchmarks, propose VPNet, and validate its effectiveness in discrete and continuous settings. Future work will explore extending VPN to real-world environments. This research is supported by the National Natural Science Foundation of China (92464204). We sincerely thank Zun Wang for providing the code used to generate discrete view images in HM3D.
%We introduce Visual Prompt Navigation (VPN), a novel paradigm that guides embodied agents using 2D top-view visual prompts without relying on natural language. We built two benchmark datasets, proposed a dedicated baseline model (VPNet), and designed effective augmentation strategies. Experiments in both discrete and continuous settings confirm the effectiveness of visual prompts. Future work includes extending VPN to real-world environments, enabling agents to interpret intuitive, hand-drawn instructions that better reflect human intentions.

\bibliography{anonymous-submission-latex-2026}

\begin{table}
\small
\noindent\begin{minipage}[t]{1\columnwidth}%
\tabcolsep=0.1cm
\centering
\begin{tabular}{c|c|cc|cc}
\toprule
\multirow{2}{*}{Split} & \multirow{2}{*}{Setting} & \multicolumn{2}{c|}{VLN}  & \multicolumn{2}{c}{VPN} \tabularnewline 
 & & Scenes & Episodes & Scenes & Episodes \tabularnewline \hline 
 % \multirow{2}{*}{Train} & Discrete & 61 & \num{4675} & 61 & \num{4638} \tabularnewline  
 % & Continuous & 61 & \num{3603} & 60 & \num{3597} \tabularnewline \hline 
 \multirow{2}{*}{Val Seen} & Discrete & 56 & \num{340} & 56 & \num{338} \tabularnewline 
 & Continuous & 53 & 259 & 52 & \num{258} \tabularnewline \hline 
 \multirow{2}{*}{Val Unseen} & Discrete & 11 & \num{783} & 11 & \num{783} \tabularnewline 
 & Continuous & 11 & 613 & 11 & \num{613} \tabularnewline \hline 
\multirow{1}{*}{Test} & Discrete & 18 & \num{1391} & 18 & \num{1391} \tabularnewline 
\bottomrule
\end{tabular}
\end{minipage}
\caption{Statistics of validation and test sets in R2R for VLN and VPN under discrete and continuous navigation settings.}
\label{data_analysis2}
\end{table}

\begin{table}[ht]
\centering
\scriptsize
\setlength{\tabcolsep}{3.2mm}
\begin{tabular}{c|ccccc}
\toprule
Fusion methods & 
\multicolumn{1}{c}{TL$\downarrow$} & \multicolumn{1}{c}{NE$\downarrow$} & \multicolumn{1}{c}{OSR$\uparrow$} & \multicolumn{1}{c}{SR$\uparrow$} & \multicolumn{1}{c}{SPL$\uparrow$}  \\ \hline
Last map & 10.04 & 0.65 & 96.04 & 95.91 & 94.11 \\
OAFC & 9.92 & \textbf{0.48} & \textbf{97.45} & \textbf{96.68} & \textbf{94.84} \\
\bottomrule
\end{tabular}
\caption{Comparison among different top-view map fusion methods. ``Last map'' refers to using only the final floor map corresponding to the agent's trajectory as the visual prompt, while ``OAFC'' denotes our proposed Order-Aware Floor Concatenation strategy mentioned in Section~\ref{v2p}.}
\label{abalation_floors}
\end{table}

\section{Appendix}

We organise the appendix mainly in four parts. Section~\ref{more_vpnet} provides more details of VPNet. More details of datasets are described in Section~\ref{more_data}. Section~\ref{more_exp} presents more experimental results of VPNet, including quantitative and qualitative analyses. Section~\ref{future} outlines our vision and potential directions for future work.

\subsection{More Details of VPNet}
\label{more_vpnet}

\noindent\textbf{Local Cross-Modal Encoder.} VPNet's local branch performs the local action prediction based on the panoramic observations at the current node, in addition to the global action prediction described in Section 4.3. Given the contextual view representations $W_i=(W_{i1},W_{i2}, \dots,W_{in})$ of the current node, the local cross-modal encoder (a standard 4-layer cross-modal transformer) models the relations between the panoramic views $W_i$ and the visual prompts $\mathcal{P}$, resulting in $W_i'$.

\noindent\textbf{Dynamic Fusion.} In discrete environments, the VPNet performs final action prediction by dynamically fusing global and local action prediction. For global action prediction, we mask the scores of previously visited nodes and the current node for $s^g_i$, resulting in ${s^{g}_i}^{\prime}$. The local branch computes $s_i^l$ on $W_i'$, which includes scores for navigable views and a special stop score. During local action prediction, masks are applied for $s^l_i$ to exclude the unnavigable views, resulting in ${s^{l}_i}^{\prime}$. Since the local action space differs from the global space, which considers all navigable nodes in $\mathcal{V}_t$, the local action scores $s_i^l$ are transformed to match the global action space. This is done by adding a backtrack score $s_b$, which is the sum of the scores of visited nodes in $\mathcal{N}(V_t)$, where $\mathcal{N}(V_t)$ denotes the neighboring nodes connected to the current node:
\begin{table}[ht]
\centering
\small
\begin{tabular}{c|ccc}
\toprule
FM / NF & 1 & 2 & 3 \\ \hline
Ground truth & 639(100\%) & 122(100\%) & 22(100\%) \\
Last map & 635(99.37\%) & 102(83.61\%) & 14(63.64\%) \\
OAFC & 634(99.22\%) & 107(87.70\%) & 16(72.73\%) \\
\bottomrule
\end{tabular}
\caption{Statistical results on the val unseen split in the discrete navigation setting for different top-view map fusion methods (FM). The columns headers represent the total number of floors (NF) that the agent is expected to traverse in the standard navigation trajectory.}
\label{floor_contrast}
\end{table}

\begin{table}[ht]
\centering
\small
\begin{tabular}{c|cc}
\toprule
VLN / VPN & Success & Failure \\
\hline
Success & 1839 & 62 \\
Failure & 432 & 16 \\
\bottomrule
\end{tabular}
\caption{Cross-success statistics between VLN and VPN on the validation unseen split in discrete environments. The rows indicate the success or failure of VLN episodes, while the columns represent the corresponding success or failure outcomes in VPN.}
\label{vln_vpn_stats}
\end{table}
\begin{equation}
{s^{l}_i}^{\prime} = 
\begin{cases} 
s_{\text{b}}, & \text{if } \mathcal{V}_{i} \in \mathcal{V}_{t} - \mathcal{N}(V_t), \\
s^{l}_i, & \text{otherwise}. \tag{11}
\end{cases}
\end{equation}
This adjustment enables the agent to navigate to unexplored nodes by retracing its steps through neighboring nodes. The final navigation score is a fusion of global and adjusted local action scores:
\begin{equation}
    s_i = \sigma_t {s^{g}_i}^{\prime} + (1 - \sigma_t) {s^{l}_i}^{\prime}                 \tag{12}
\end{equation}
where $\sigma_t$ is a learnable scalar for fusion.

\subsection{More Details of Datasets}
\label{more_data}

Table~\ref{data_analysis2} presents the split-wise statistics for the validation and test sets. For evaluation, we use the validation splits of R2R-VP and R2R-CE-VP, corresponding to the discrete and continuous environments. Note that in the R2R-VP validation split, we exclude trajectories associated with low-quality top-view map reconstructions, resulting in a slight reduction in the number of scenes and episodes. The VPN test set is derived from ScaleVLN's predictions on the R2R test split by extracting the shortest paths between start and goal viewpoints.

\subsection{More Experimental Results}
\label{more_exp}

\noindent\textbf{Abalation Study on Visual Prompt Map Fusion.} As shown in Table~\ref{abalation_floors}, ``OAFC'' consistently outperforms ``Last map'' across all metrics. Unlike ``Last map'', which uses only the final floor map, ``OAFC'' preserves the full sequence of visual prompt maps and encodes their temporal order for more informative fusion. Furthermore, Table~\ref{floor_contrast} shows that ``OAFC'' offers greater benefits in multi-floor scenarios. For two-floor trajectories, ``OAFC'' achieves 87.70\% compared to 83.61\% with ``Last map''; the gap increases to 72.73\% vs. 63.64\% in three-floor cases. These results suggest ``OAFC'' is more effective in capturing multi-floor spatial information under complex navigation settings.

\begin{figure*}[t]
    \centering
    \begin{subfigure}[t]{\textwidth}
        \centering
        \includegraphics[width=\textwidth]{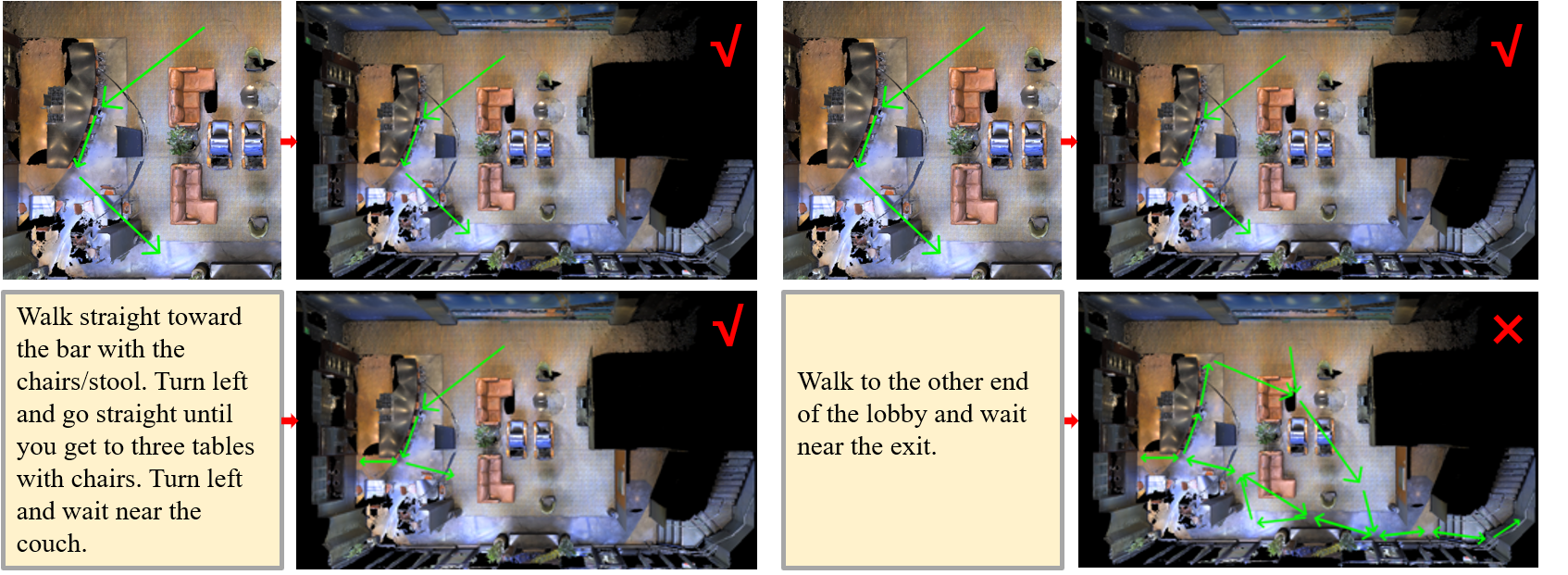}
        \caption{Predicted trajectories by VPNet (top) and DUET (bottom) for the same navigation episode. The two columns of subfigures correspond to sampled cases from the first column (i.e., VPN Success vs. VLN Success and Failure) in Table~\ref{vln_vpn_stats}.}
        \label{sub_ps}
    \end{subfigure}
    
    \begin{subfigure}[t]{\textwidth}
        \centering
        \includegraphics[width=\textwidth]{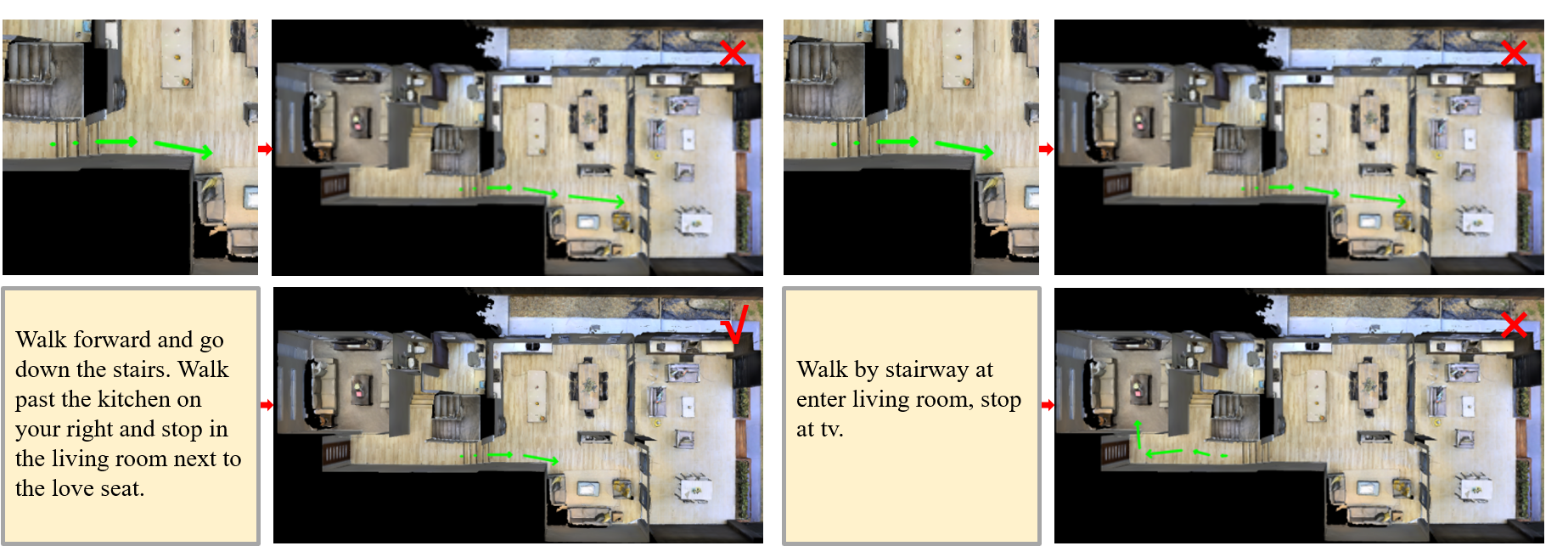}
        \caption{Predicted trajectories by VPNet (top) and DUET (bottom) for the same navigation episode. The two columns of subfigures correspond to sampled cases from the second column (i.e., VPN Failure vs. VLN Success and Failure) in Table~\ref{vln_vpn_stats}.}
        \label{sub_pf}
    \end{subfigure}
    
    \caption{Comparison of predicted trajectories by VPNet and DUET on the validation unseen split of R2R/R2R-VP. Both models trained with ``R2R+PRE+SCA''. Each subfigure consists of four panels labeled (1)–(4) from left to right, top to bottom.}
    \label{vpnet_vs_duet}
\end{figure*}

\noindent\textbf{Cross-Success Statistics.} Table~\ref{vln_vpn_stats} summarizes the cross-success rates between VLN (DUET) and VPN (VPNet) on the validation unseen split. Each entry in VPN is counted three times, as each trajectory in VLN is paired with three instructions. Both models are trained on the ``R2R+PRE+SCA'' training datasets. The results highlight the effectiveness of using visual prompts as navigation instructions, as VPNet succeeds in many cases where DUET fails.

\noindent\textbf{Qualitative Examples.} As Figure~\ref{sub_ps} illustrates, VPNet successfully predicts a trajectory that perfectly aligns with the ground truth, whereas DUET's performance varies notably with the level of linguistic granularity. Specifically, DUET succeeds when processing the more detailed instruction in panel 3, but fails to reach the target location when given the less specific guidance in panel 4. This observation is consistent with the dilemma discussed in Section 1 regarding human-robot interaction via natural language: while detailed instructions can enhance navigation accuracy, they often introduce verbosity that may negatively affect user experience. Furthermore, in panels 1 and 3, although both models eventually reach the destination, VPNet outperforms DUET in terms of trajectory efficiency by planning a shorter trajectory.

Figure~\ref{sub_pf} presents a failure case for VPNet. Notably, although VPNet fails to reach the correct destination, it still passes by it, indicating a relatively minor deviation. In contrast, DUET only succeeds in panel 3 when given a detailed instruction, yet entirely fails in panel 4 by steering in the opposite direction when given a vague one. This not only further confirms the aforementioned dilemma regarding human-robot interaction via natural language but also highlights the effectiveness of visual prompts as instructions to guide robot navigation.

\subsection{Discussion and Future Work}
\label{future}

We envision a future where top-view maps, acquired through aerial photography by drones or generated via multi-view 3D scene reconstruction, serve as the foundation for human-robot interaction. Based on such maps displayed on touchscreen interfaces (e.g., tablets), users could easily provide visual prompt instructions by simply tapping on a sequence of locations. The system would then automatically connect these points to form navigation cues, eliminating the need to manually draw arrows or paths and significantly lowering the barrier to instruction creation.

In addition, we plan to further explore how robots can navigate effectively using even more simplified visual prompts, and to develop models that are better suited for visual prompt navigation.

\end{document}